\documentclass[sigconf]{acmart}
\AtBeginDocument{%
  }

\copyrightyear{2026}
\acmYear{2026}
\setcopyright{cc}
\setcctype{by}
\acmConference[MM '26]{Proceedings of the 34th ACM International Conference on Multimedia}{November 10--14, 2026}{Rio de Janeiro, Brazil}
\acmBooktitle{Proceedings of the 34th ACM International Conference on Multimedia (MM '26), November 10--14, 2026, Rio de Janeiro, Brazil}
\acmDOI{10.1145/3767308.3836073}
\acmISBN{979-8-4007-2213-4/2026/11}

\usepackage{multirow}
\usepackage{cleveref}
\usepackage[table]{xcolor}

\begin{document}

\title{GraphCAD: Geometry-Aware Diffusion on Edge-Adjacency Graphs for CAD Generation}

\author{Wenhao Zheng}
\orcid{0000-0002-4203-0852}
\affiliation{
 \institution{College of Computer Science, Sichuan University} 
 \city{Chengdu} 
 \postcode{610065}
 \country{China}
 }
 \email{zhengwenhao@stu.scu.edu.cn}
 \additionalaffiliation{\institution{Engineering Research Center of Machine Learning and Industry Intelligence, Ministry of Education, Chengdu, China}}

\author{Xianggen Liu} \authornotemark[1] \authornotemark[2]
\orcid{0000-0001-7340-4602}
\affiliation{
 \institution{College of Computer Science, Sichuan University} 
 \city{Chengdu} \country{China}
 }
\email{liuxianggen@scu.edu.cn}

\author{Li Yue}
\orcid{0000-0002-8519-2114}
\affiliation{
 \institution{Dongfang Electric (Chengdu) Innovation Research Co., Ltd.} 
 \city{Chengdu} \country{China}
 }
\email{liy1383@dongfang.com}

\author{Jiancheng Shi}
\orcid{0000-0002-3234-7872}
\affiliation{
 \institution{Southwest China Research Institute of Electronic Equipment} 
 \city{Chengdu} \country{China}
 }
  \email{shijcbit@163.com}

\author{Jiancheng Lv} \authornotemark[1] 
\orcid{0000-0001-6551-3884}
\authornote{Corresponding authors.}
\affiliation{
 \institution{College of Computer Science, Sichuan University} 
 \city{Chengdu} \country{China}
 }
\email{lvjiancheng@scu.edu.cn}

\renewcommand{\shortauthors}{Wenhao Zheng, Xianggen Liu, Li Yue, Jiancheng Shi, \& Jiancheng Lv}

\begin{abstract}
Generative modeling for Computer-Aided Design (CAD) remains fundamentally challenged by the need to capture the intricate interplay between discrete topological connectivity and precise spatial geometry inherent in Boundary Representations (B-reps). Existing generative approaches largely formulate CAD construction as a sequence modeling problem, but consequently lack an explicit topological inductive bias. In this work, we present \textbf{GraphCAD}, a geometry-aware generative framework that centers modeling on the Edge-Adjacency Graph (EAG), a novel graph-based CAD representation tailored for generative tasks. GraphCAD adopts a hybrid graph diffusion formulation to jointly model the discrete and continuous components of EAG and introduces a geometry-aware denoising mechanism that ``\textit{renders}'' node attributes into spatial point sets. These point sets are encoded by an SE(3)-equivariant backbone and fused into the node representations, thereby grounding the generative process in physical geometry. To support this study, we construct EAGBench, a large-scale dataset of engineering CAD models. Extensive experiments show that GraphCAD can generate complex CAD models with more than 50 faces and achieves state-of-the-art performance in geometric fidelity and distributional alignment compared with sequence-based baselines.
\end{abstract}

\ccsdesc[500]{Computing methodologies~Shape representations}
\ccsdesc[500]{Computing methodologies~Reconstruction}

\keywords{Computer-Aided Design (CAD), Graph Diffusion, Representation Learning}



\maketitle

\section{Introduction}

The foundation of modern engineering and manufacturing depends critically on Computer-Aided Design (CAD) models for the rigorous specification of physical objects~\cite{shivegowda2022review,deng2025mamtiff,su2026equicad}. Although automating the synthesis of such models is highly desirable for accelerating design workflows, the inherently hybrid structure of CAD data makes this problem particularly challenging~\cite{patel2006automatic,chen2025img2cad,li2025stitch}. In practice, CAD data is most commonly represented in the Boundary Representation (B-rep) format~\cite{jayaraman2023solidgen}, which defines 3D models through parameterized faces, edges, and vertices, jointly encoding the geometric and topological information necessary for a complete description of the shape~\cite{jones2023b,liu2025hola,lee2025brepdiff}. As a result, directly synthesizing B-reps remains notoriously difficult, since it requires modeling continuous geometric parameters and discrete topological connectivity in a unified manner~\cite{guo2022complexgen,liu2024point2cad,liu2024split}.

\begin{figure}[t]
  \centering
  \includegraphics[width=0.8\linewidth]{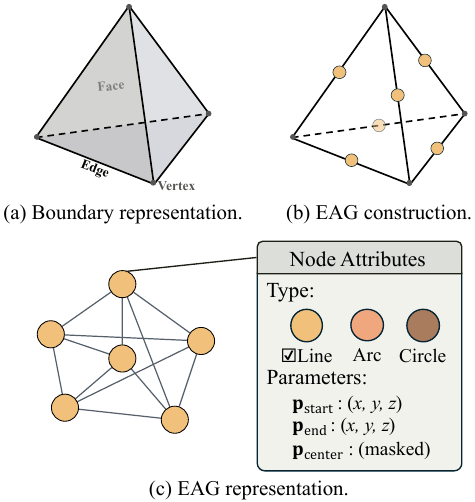}
  \vspace{-7pt}
\caption{
Construction of the Edge-Adjacency Graph (EAG) representation.
(a) A Boundary Representation (B-rep) of a CAD solid.
(b) Construction of the Edge-Adjacency Graph, where each graph node corresponds to a topological edge in the B-rep and graph edges indicate shared B-rep vertices.
(c) The EAG representation with node attributes, where each node encodes the curve type together with compact analytic geometric parameters.
}
  \label{fig:data}
  \vspace{-10pt}
\end{figure}

Prevailing approaches typically model B-reps either as sequences of commands~\cite{li2022free2cad,ren2022extrudenet,khan2024cad} or as Constructive Solid Geometry (CSG) programs~\cite{ren2021csg,sharma2018csgnet,yu2023d}. While these paradigms are intuitive, they often struggle to preserve strict topological integrity~\cite{xu2024brepgen,li2025dtgbrepgen}. In sequence-based generation, for example, a small error in an early operation can propagate through the entire construction process, ultimately producing disjoint loops, non-manifold edges, or entirely invalid geometries. More recently, methods such as BrepGiff~\cite{guo2025brepgiff} and GraphBRep~\cite{lai2026graphbrep} have sought to mitigate these topological issues by abstracting B-reps into Face-Adjacency Graphs~\cite{jayaraman2021uv}. By leveraging the structural advantages of graph representations, these methods have shown promising results. However, because they typically rely on UV-domain sampling to approximate geometry and do not explicitly anchor generation to a 1D wireframe scaffold, they remain limited in enforcing the precise boundary constraints required for high-fidelity industrial CAD models.

To address these limitations, we revisit prevailing graph formulations and introduce GraphCAD. In contrast to prior approaches based on Face-Adjacency Graphs, where nodes correspond to complex trimmed surfaces, our method adopts a new graph construction paradigm centered on 1D wireframe edges. Specifically, we propose the \textit{Edge-Adjacency Graph (EAG)}, which redefines the fundamental graph unit from faces to edges. Under this formulation, each node represents a topological edge augmented with hybrid attributes, including curve types and spatial parameters.

To synthesize the proposed EAG representation, GraphCAD employs a geometry-aware hybrid graph diffusion framework. To accommodate the heterogeneous nature of EAG, we formulate a \textit{hybrid diffusion process} that jointly models the distribution of discrete topological attributes and continuous geometric parameters. Unlike standard graph diffusion models, which treat node attributes as independent numerical vectors, our method explicitly interprets them as spatial geometry. During diffusion, we ``\textit{render}'' the numerical attributes associated with topological B-rep edges into 3D curves, sample local point sets from these curves, and encode them with an SE(3)-equivariant backbone. This design grounds abstract parameters in their physical realization, enabling the model to capture the intricate interplay between topological connectivity and geometric fidelity inherent to the EAG representation. To validate our approach, we introduce {EAGBench}, a large-scale unified dataset constructed by aggregating and standardizing component-level B-reps from existing repositories such as DeepCAD~\cite{wu2021deepcad} and the Fusion 360 Gallery~\cite{willis2021fusion}.

Our contributions can be summarized as follows:
\begin{itemize}
    \item We introduce the Edge-Adjacency Graph (EAG), a novel graph representation of Boundary Representations (B-reps) specifically designed for CAD generation.

    \item We present GraphCAD, a geometry-aware graph diffusion framework for CAD generation. Its SE(3)-equivariant denoising network produces predictions that transform consistently under rigid transformations while effectively modeling the hybrid distribution of EAG data.
    
    \item We propose a novel geometry rendering mechanism that bridges abstract graph attributes and physical shape realization. By ``\textit{rendering}'' node attributes into 3D curves and sampling spatial point sets, this mechanism enables the graph neural network to better perceive and encode the underlying CAD geometry.
    
    \item We construct {EAGBench}, a large-scale unified benchmark built upon the EAG representation. Extensive experiments on EAGBench show that GraphCAD achieves state-of-the-art performance in geometric fidelity and distributional alignment, while maintaining highly competitive topological validity compared with leading sequence-based baselines.
\end{itemize}
\section{Related Work}
\label{sec:related_work}

\subsection{Generative CAD Modeling}

\begin{figure*}[ht]
  \centering
  \includegraphics[width=\textwidth]{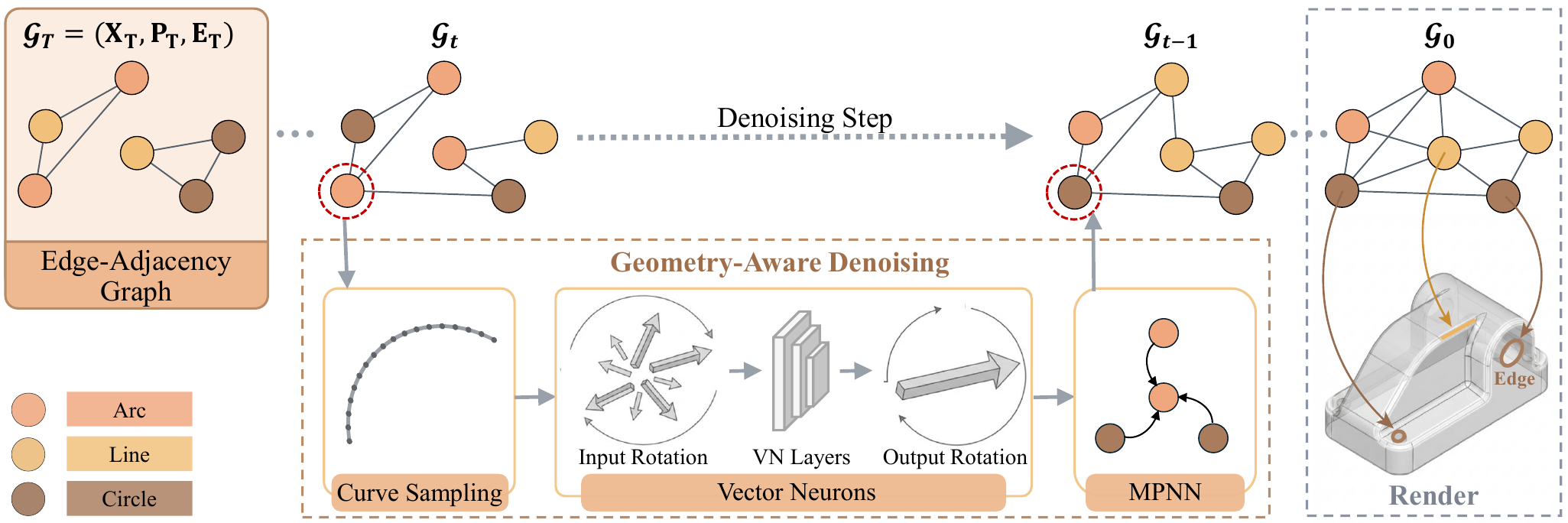}
  \vspace{-18pt}
    \caption{
    Overview of the GraphCAD framework.
    \textbf{Top:} The generative pipeline. The model synthesizes CAD models by iteratively denoising the Edge-Adjacency Graph (EAG), our proposed topological representation of B-rep geometry.
    \textbf{Bottom:} The Geometry-Aware Denoising module. Unlike standard graph-based methods that treat node attributes as abstract vectors, our approach explicitly ``\textit{renders}'' these attributes into spatial point sets through ``\textit{Curve Sampling}''. The sampled point sets are then processed by an SE(3)-equivariant Vector Neuron backbone to extract rotation-robust geometric embeddings.
    }
  \label{fig:geometry_aware_diffusion}
  \vspace{-10pt}
\end{figure*}

The generation of CAD models has attracted increasing attention in recent years~\cite{xu2023hierarchical,rukhovich2025cad,casey2025aligning}. Early approaches often represented CAD shapes using Constructive Solid Geometry (CSG)~\cite{yin2020topologically,yu2022capri,mews2025dontmeshmegenerating}. Although theoretically expressive, CSG-based methods are generally regarded as overly complex and insufficiently flexible for typical industrial design scenarios~\cite{andujar2022solid}. As a result, much of the recent literature has shifted toward feature-based modeling paradigms in which shapes are represented as sequences of commands~\cite{wang2025texttocad,yuan2025cadeditor,li2025cad,xu2022skexgen}. Recent methods such as Text2CAD~\cite{khan2024text2cad} and CADDreamer~\cite{li2025caddreamer} have substantially improved controllability and geometric fidelity within this paradigm. Nevertheless, these approaches still rely on sequential or implicit formulations. Because they primarily encode command histories rather than explicit geometric-topological structure, they offer limited direct access to topological and geometric information and therefore lack an explicit topological inductive bias. Consequently, a critical gap remains for generative methods that can directly synthesize robust CAD representations.

\subsection{Representation Learning for B-reps}
Boundary Representation (B-rep) defines a solid through geometric primitives, such as parametric curves and surfaces, together with topological entities including vertices, edges, and faces~\cite{lou2023brep}. While discrete representations, such as point clouds~\cite{qi2017pointnet}, voxels~\cite{wu20153d}, meshes~\cite{hanocka2019meshcnn}, and sequences~\cite{ganin2021computer}, have been extensively studied, they lack the exact geometric precision required for engineering design. To bridge the gap between deep learning and precise CAD modeling, prior work has sought to encode B-reps into learnable graph structures~\cite{colligan2022hierarchical,xia2024machining,dai2025brepformer}. UV-Net~\cite{jayaraman2021uv} pioneered the representation of B-reps as Face-Adjacency Graphs (FAGs), in which nodes are associated with 2D surface features extracted from UV grids. This face-centric formulation was later adopted and extended by generative methods such as BrepGiff~\cite{guo2025brepgiff} and GraphBRep~\cite{lai2026graphbrep}, which leverage graph structures to preserve complex topological relationships during generation. However, because these methods typically represent nodes as complex trimmed surfaces and decompose generation into multiple submodules, they often suffer from computational redundancy, error accumulation across stages, and substantial difficulty in enforcing precise 1D boundary constraints.

\subsection{Graph Diffusion and Equivariance}
Diffusion models~\cite{ho2020denoising,dhariwal2021diffusion} have achieved state-of-the-art performance across a wide range of generative tasks. In the graph domain, methods such as GDSS~\cite{jo2022score} and DiGress~\cite{vignac2023digress} have successfully extended diffusion modeling to continuous and discrete graph structures, respectively. However, applying these generic frameworks to CAD requires addressing domain-specific engineering constraints, particularly the physical symmetries of 3D space. SE(3)-equivariant networks~\cite{fuchs2020se3} have demonstrated strong effectiveness in applications such as molecular dynamics by ensuring robustness to rotations. In this work, we bring these advances into the CAD domain. Building upon discrete graph diffusion, we develop an SE(3)-equivariant hybrid model specifically tailored to the Edge-Adjacency Graph. This design enables the generation of rigorous engineering geometries with invariant likelihoods, a property that remains largely overlooked in existing graph generative models.

\section{Method}
\label{sec:method}

We present a novel generative framework for CAD models (see Figure~\ref{fig:geometry_aware_diffusion}). To introduce a stronger topological inductive bias in sequence-based methods, we shift the representation paradigm from parametric CAD sequences to the {Edge-Adjacency Graph (EAG)}. This formulation introduces strong structural inductive biases that naturally promote topological integrity and, in turn, improve generation quality. Based on this representation, we develop a geometry-aware hybrid graph diffusion model that jointly generates both structural connectivity and precise 3D geometry, treating them as physical spatial entities rather than independent abstract variables.

\begin{figure*}[t]
  \centering
  \includegraphics[width=1.0\linewidth]{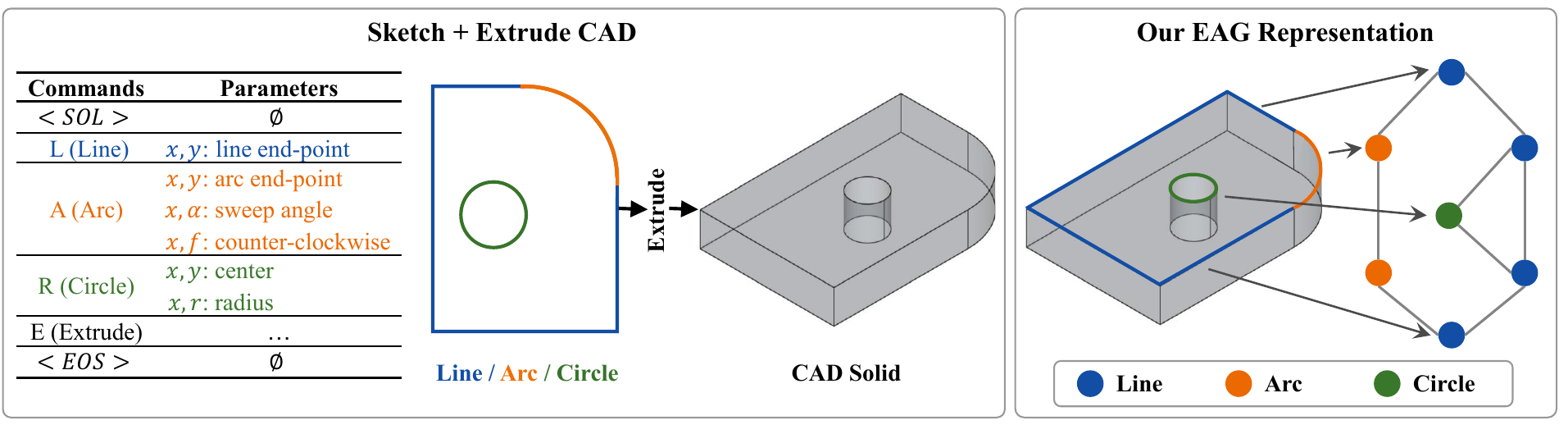}
  \vspace{-16pt}
\caption{
Relationship between sketch-and-extrude CAD and the EAG representation. EAG encodes the resulting line, arc, and circle boundary primitives as typed nodes, while graph edges capture their topological adjacency.
}
  \label{fig:eag_sketch_extrude}
  \vspace{-10pt}
\end{figure*}

\subsection{Edge-Adjacency Graph Representation}
\label{subsec:representation}

Conventional graph-based representations of B-reps typically adopt a Face-Adjacency Graph, in which nodes correspond to parametric surfaces. Although effective for discriminative tasks, the intrinsic complexity of these high-dimensional surface features makes learning their joint distribution computationally prohibitive for generative modeling. To address this limitation, we identify the \textit{wireframe} as the primary structural skeleton of a solid. Accordingly, as illustrated in Figure~\ref{fig:data}, we shift the graph formulation from faces to edges.

Formally, given a Boundary Representation $\mathcal{B}$, we construct an undirected graph $\mathcal{G} = (\mathcal{V}, \mathcal{E})$.

\begin{definition}[\textbf{Edge Nodes}]
Let $\mathcal{V} = \{v_1, \dots, v_N\}$ be the set of nodes, where each node $v_i$ corresponds to a unique topological edge $e^{topo}_i$ in $\mathcal{B}$. Associated with each node $v_i$ is a compact geometric feature vector $\mathbf{h}_i \in \mathbb{R}^{12}$.
\end{definition}

In our specific implementation, we adopt a concise analytical parameterization. The feature vector is structured as a concatenation of primitive attributes: 
\begin{equation}
  \label{eq:node_attributes}
  \mathbf{h}_i = [\mathbf{x}_{\text{type}}, \mathbf{p}_{\text{start}}, \mathbf{p}_{\text{end}}, \mathbf{p}_{\text{center}}],  
\end{equation}
where $\mathbf{x}_{\text{type}} \in \{0,1\}^3$ is a one-hot encoding of the curve type. In our representation, edge primitives are categorized into three analytical types: \textit{Line}, \textit{Arc}, and \textit{Circle}. For \textit{Spline} curves, we adopt a piecewise-arc approximation and represent each spline as a sequence of arc segments in the graph. Each $\mathbf{p} \in \mathbb{R}^3$ denotes a 3D spatial point.  For \textit{Line} primitives, $\mathbf{p}_{\text{center}}$ is masked. For \textit{Circle} primitives, $\mathbf{p}_{\text{start}}$ and $\mathbf{p}_{\text{end}}$ are sampled on the circumference such that the chord connecting them does not pass through $\mathbf{p}_{\text{center}}$. This geometric constraint ensures that the three points are non-collinear, thereby uniquely determining the circle's 3D plane and curvature. As illustrated in Figure~\ref{fig:eag_sketch_extrude}, these primitives capture the line, arc, and circle boundaries produced by common sketch-and-extrude operations, including solids with curved boundaries. The corresponding surfaces are recovered during B-rep reconstruction.

\begin{definition}[\textbf{Topological Connectivity}]
The set of graph edges $\mathcal{E}$ defines the connectivity between physical CAD edges. We define an edge $(v_i, v_j) \in \mathcal{E}$ if the corresponding CAD edges $e^{topo}_i$ and $e^{topo}_j$ share a common vertex in the B-rep data structure.
\end{definition}

\subsection{Hybrid Graph Diffusion}
\label{subsec:diffusion}

To effectively model the distribution of this graph representation, we introduce a \textit{hybrid graph diffusion model} $\mathcal{M}_{\theta}$ for EAG generation. Specifically, we extend the discrete graph diffusion framework to a hybrid state space that jointly captures the distribution of discrete semantic attributes and continuous geometric parameters (see Appendix B for the detailed mathematical derivation).

\textbf{Data Formulation.}
Let $G = (\mathcal{V}, \mathcal{E})$ be the Edge-Adjacency Graph with $n$ nodes. Unlike standard discrete diffusion models that operate solely on categorical data, our node attributes lie in a product space of discrete semantics and continuous geometry, represented by spaces $\mathcal{X}$ and $\mathcal{P}$, respectively. We use $\mathbf{x}_i \in \{0,1\}^{d_x}$ to denote the one-hot encoded semantic type (Line, Arc, Circle) of node $i$, where $d_x=3$, organized in a matrix $\mathbf{X} \in \mathbb{R}^{n \times d_x}$. Similarly, we use $\mathbf{p}_i \in \mathbb{R}^{d_p}$ to denote the continuous geometric parameters (e.g., coordinates, radii), organized in a parameter matrix $\mathbf{P} \in \mathbb{R}^{n \times d_p}$, where $d_p=9$. The tensor $\mathbf{E}_t \in \mathbb{R}^{n \times n \times d_e}$ is the one-hot encoding of the edge set $\mathcal{E}$. The complete state of the graph at time $t$ is thus denoted by the tuple $\mathcal{G}_t = (\mathbf{X}_t, \mathbf{P}_t, \mathbf{E}_t)$.

\textbf{Forward Process.}
We define a forward diffusion process $q$ that progressively corrupts the clean graph $\mathcal{G}_0$ over $T$ timesteps. Crucially, the noise is applied independently to the discrete, continuous, and topological components. The joint transition distribution factorizes as:
\begin{equation}
\begin{aligned}
q(\mathcal{G}_t \mid \mathcal{G}_{t-1})
&=
q_{\text{disc}}(\mathbf{X}_t \mid \mathbf{X}_{t-1})
\\
&\quad \times\,
q_{\text{cont}}(\mathbf{P}_t \mid \mathbf{P}_{t-1})
\times
q_{\text{edge}}(\mathbf{E}_t \mid \mathbf{E}_{t-1}).
\end{aligned}
\end{equation}

For the \textit{discrete components} (semantics $\mathbf{X}_t$ and topology $\mathbf{E}_t$), we adopt the matrix transition formulation from DiGress. The noise is defined by a sequence of Markov transition matrices $\mathbf{Q}_t$. The forward transition applies the noise independently to each node (or edge), sampling from a categorical distribution:
\begin{equation}
    \mathbf{X}_t \sim \mathcal{C}at(\mathbf{X}_{t-1} \mathbf{Q}_t).
\end{equation}

A key property of this Markov chain is that the marginal distribution at any arbitrary timestep $t$ can be computed in closed form as $\mathbf{X}_t \sim \mathcal{C}at(\mathbf{X}_0 \overline{\mathbf{Q}}_t)$, where $\overline{\mathbf{Q}}_t = \mathbf{Q}_1 \dots \mathbf{Q}_t$.

For the \textit{continuous component} $\mathbf{P}$, we employ a standard Gaussian diffusion process with a variance schedule $\beta_t$. The transition kernel for the parameter matrix is defined as:
\begin{equation}
    \mathbf{P}_t = \sqrt{1-\beta_t}\mathbf{P}_{t-1} + \sqrt{\beta_t}\boldsymbol{\mathcal{E}}, \quad \boldsymbol{\mathcal{E}} \sim \mathcal{N}(\mathbf{0}, \mathbf{I}),
\end{equation}
where $\boldsymbol{\mathcal{E}} \in \mathbb{R}^{n \times d_p}$ is the sampled Gaussian noise matrix.

\textbf{Reverse Process.}
During generation, the reverse transition is conditionally factorized as
\begin{equation}
\begin{aligned}
p_{\theta}(\mathcal{G}_{t-1}\mid\mathcal{G}_t)
={}&q(\mathbf{P}_{t-1}\mid\mathbf{P}_t,\boldsymbol{\epsilon}_{\theta}) \\
&\times q(\mathbf{X}_{t-1}\mid\mathbf{X}_t,\hat{\mathbf{X}}_0)
q(\mathbf{E}_{t-1}\mid\mathbf{E}_t,\hat{\mathbf{E}}_0).
\end{aligned}
\end{equation}
Here,
$(\hat{\mathbf{X}}_0,\hat{\mathbf{E}}_0,
\boldsymbol{\epsilon}_{\theta})
=M_{\theta}(\mathcal{G}_t,t)$.
Although conditionally factorized for tractable sampling, the three components remain coupled through the shared denoising network $M_{\theta}(\mathcal{G}_t,t)$, which processes the complete noisy graph and jointly predicts $(\hat{\mathbf{X}}_0,\hat{\mathbf{E}}_0,
\boldsymbol{\epsilon}_{\theta})$.

\subsection{Geometry-Aware Denoising}
\label{subsec:network}

To fully leverage the geometric priors inherent in the EAG representation, we design a geometry-aware denoising network $\mathcal{M}_\theta(\mathcal{G}_t, t)$ that explicitly grounds latent attributes in their physical realization. Standard graph diffusion architectures typically treat node features as abstract numerical vectors. In our framework, however, these attributes encode explicit spatial definitions that are essential for valid construction. To effectively capture this geometric structure, we adopt a modular architecture composed of three stages: \textit{Geometric Feature Extraction}, \textit{Equivariant Backbone}, and \textit{Hybrid Prediction Heads}.

\textbf{Geometric Feature Extraction.}
To provide grounded spatial guidance, we explicitly ``\textit{render}'' the noisy latent parameters into a point-based representation before feeding them to the backbone. For every node $v_i$ at timestep $t$:

\begin{itemize}
    \item \textbf{B\'{e}zier Instantiation:} We interpret the noisy parameters $\mathbf{p}_{i,t}$ as control points for {Quadratic B\'{e}zier curves} conditioned on $\mathbf{x}_{i,t}$. This relaxation robustly generates valid, differentiable 3D loci even when intermediate states violate strict geometric invariants (e.g., unequal radii).

    \item \textbf{Point Sampling:} We uniformly sample $K=100$ points along this reconstructed noisy curve to generate a local discrete point set:
    \begin{equation}
        \mathcal{P}_{i,t} = \{\mathbf{u}_k\}_{k=1}^{K},
    \end{equation}
    where $\mathcal{P}_{i,t}$ comprises $K$ sampled coordinates $\mathbf{u}_k \in \mathbb{R}^3$ for node $v_i$ at timestep $t$.

    \item \textbf{Geometric Encoding:} To ensure SE(3) equivariance (see Sec.~\ref{subsec:equivariance}), we process $\mathcal{P}_{i,t}$ using a Vector Neuron~\cite{deng2021vector} encoder $\psi_{\text{vn}}$. This module maps the point set to a rotation-equivariant embedding $\mathbf{g}_{i,t}=\psi_{\text{vn}}(\mathcal{P}_{i,t}) \in \mathbb{R}^{d_g \times 3}$, comprising $d_g$ distinct 3D vector features. This embedding satisfies
        \begin{equation}
        \label{eq:se3}
            \psi_{\text{vn}}(\mathbf{R} \cdot \mathcal{P}_{i,t}) = \mathbf{R} \cdot \psi_{\text{vn}}(\mathcal{P}_{i,t}),
        \end{equation}
        for any rotation matrix $\mathbf{R} \in SO(3)$, where the rotation is applied to the spatial dimension of each feature vector.
\end{itemize}

\textbf{Equivariant Backbone.}
The backbone $\Phi_{\text{backbone}}$ processes the node attributes $\mathbf{h}_i$ (defined in Eq.~\ref{eq:node_attributes}) by partitioning them into two distinct feature streams:
a \emph{scalar stream} $\mathbf{s}_i$ containing rotation-invariant features (specifically, one-hot node types), and a \emph{vector stream} $\mathbf{V}_i \in \mathbb{R}^{(d_g+3) \times 3}$ formed by concatenating the geometric parameters $\mathbf{p}_i$ with the rendered point-set embeddings $\mathbf{g}_i$.
Inside the network, these streams are updated in parallel: $\mathbf{s}_i$ via standard MLPs and $\mathbf{V}_i$ via SE(3)-equivariant operations. This dual-stream design preserves rotational equivariance within the denoising backbone while maintaining geometric consistency.

\textbf{Hybrid Prediction Heads.}
The network outputs are generated by hybrid prediction heads $\phi_{\text{head}}$: classifiers for the discrete node types and edge topology, and a regressor for the continuous parameters. Specifically, we train the model to predict the \emph{clean start state} for both discrete structures ($\mathbf{X}_0, \mathbf{E}_0$) and the \emph{noise term} $\boldsymbol{\epsilon}$ for the continuous parameters via a hybrid loss function:
\begin{equation}
\mathcal{L} = \mathcal{L}_{\text{CE}}(\mathbf{X}_0, \hat{\mathbf{X}}_\theta) + \mathcal{L}_{\text{CE}}(\mathbf{E}_0, \hat{\mathbf{E}}_\theta) + \lambda \| \mathbf{Mask} \odot (\boldsymbol{\epsilon} - \boldsymbol{\epsilon}_\theta) \|^2.
\end{equation}
Here, the first two terms represent the cross-entropy losses for node semantic types and edge connectivity. The third term supervises the continuous geometry, weighted by $\lambda$. Crucially, $\mathbf{Mask}$ is a semantic-aware binary mask derived from the ground truth types $\mathbf{X}_0$. It zeros out the loss for geometric attributes that are undefined for a specific curve type (e.g., the center parameter for a \textit{Line}), preventing the model from hallucinating gradients for non-existent geometry.

\subsection{SE(3) Equivariance}
\label{subsec:equivariance}

We focus on the SE(3)-equivariance of the denoising network
$M_{\theta}(\mathcal{G}_t,t)$. Let $\mathbb{G}$ denote the space of
Edge-Adjacency Graphs.

\begin{definition}[\textbf{SE(3) Group Action}]
\label{def:se3_action}
We define the action of a rigid body transformation $r = (\mathbf{R}, \mathbf{t}) \in SE(3)$ on an edge-adjacency graph $G_t \in \mathcal{G}$, denoted as $r \circ G_t$. Formally, let $\Gamma: \mathcal{G} \to \mathcal{B}$ be the mapping from EAGs to solid B-reps. The action is defined by the commutative property via the underlying geometric coordinates:
\begin{equation}
r \circ G_t \triangleq \Gamma^{-1}(r \cdot \Gamma(G_t)).
\end{equation}

This formulation ensures that transforming the abstract graph representation corresponds to rotating and translating the underlying CAD geometry. By handling translation through input canonicalization and rotation through an SO(3)-equivariant backbone, the denoising network satisfies the following equivariance property.

\end{definition}

\begin{table*}[t]
\centering
\caption{
Quantitative evaluation of \textit{unconditional generation} across two benchmarks.
We measure distributional alignment between generated and reference Edge-Adjacency Graphs (EAGs) using maximum mean discrepancy over node degree (\emph{Deg}), clustering coefficient (\emph{Clus}), and orbit counts (\emph{Orb}).
Generative quality is further evaluated in terms of \emph{Validity}, \emph{Uniqueness}, and \emph{Novelty}.
Best results are shown in \textbf{bold}.
}
\vspace{-7pt}
\label{tab:graph_metrics}
\begin{tabular}{lcccccc}
\toprule
\textbf{Model} & \textbf{Deg} $\downarrow$ & \textbf{Clus} $\downarrow$ & \textbf{Orb} $\downarrow$ & \textbf{Validity} $\uparrow$ & \textbf{Uniqueness} $\uparrow$ & \textbf{Novelty} $\uparrow$  \\
\bottomrule
\multicolumn{7}{c}{\textbf{EAGBench-DC}} \\
\toprule
GraphNVP & 0.836 & 0.749 & 1.634 & 52.79 & 89.79 & 81.96 \\
GDSS     & 0.659 & 0.721 & 1.310 & 80.08 & 85.21 & 79.50 \\
ConGress & 0.313 & 0.399 & 0.559 & 71.97 & 91.20 & 85.18 \\
DiGress  & 0.242 & 0.383 & 0.183 & 95.31 & 93.42 & 89.34 \\
\rowcolor{gray!15}
\textbf{GraphCAD~(Ours)} & \textbf{0.176} & \textbf{0.248} & \textbf{0.120} & \textbf{98.05} & \textbf{95.76} & \textbf{92.21} \\
\bottomrule
\multicolumn{7}{c}{\textbf{EAGBench-FS}} \\
\toprule
GraphNVP & 0.202 & 0.402 & 0.172 & 30.03 & 88.26 & 83.14 \\
GDSS     & 0.199 & 0.297 & 0.159 & 54.08 & 87.19 & 84.25 \\
ConGress & 0.178 & 0.244 & 0.131 & 53.12 & 93.27 & 92.13 \\
DiGress  & 0.169 & 0.239 & 0.127 & 77.45 & 95.16 & 95.82 \\
\rowcolor{gray!15}
\textbf{GraphCAD~(Ours)} & \textbf{0.162} & \textbf{0.233} & \textbf{0.114} & \textbf{96.58} & \textbf{99.12} & \textbf{98.96} \\
\bottomrule
\end{tabular}
\end{table*}

\begin{figure}[t]
    \centering
    \includegraphics[width=1\linewidth]{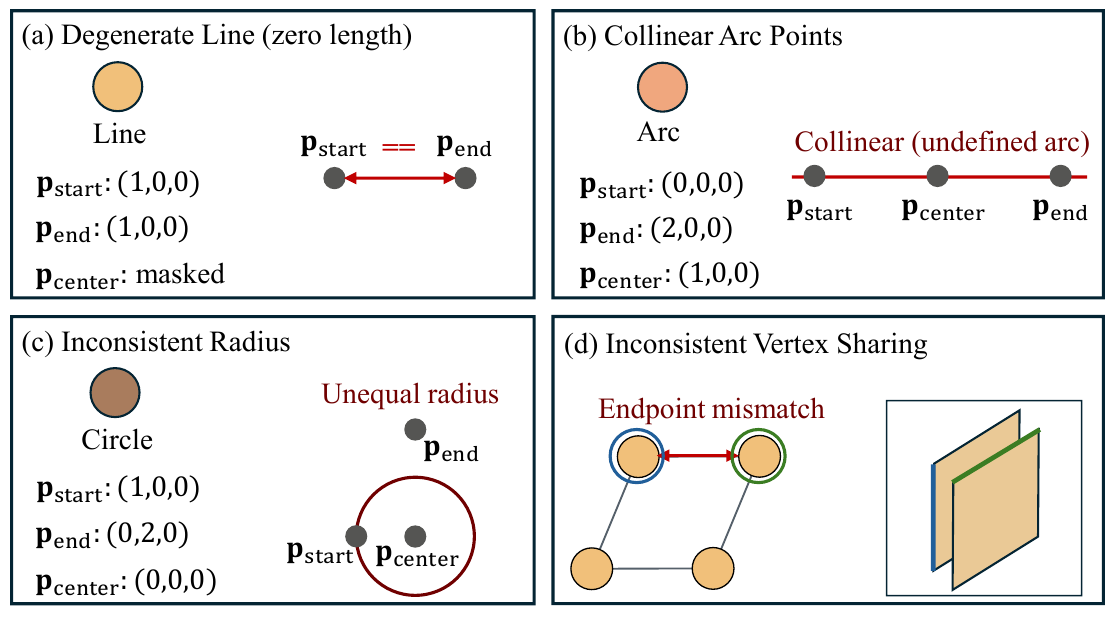}
    \vspace{-18pt}
    \caption{Visual categorization of the primary failure modes encountered during EAG-to-B-rep reconstruction.}
    \label{fig:failure_modes}
    \vspace{-10pt}
\end{figure}

\begin{theorem}[\textbf{SE(3) Equivariance}]
\label{thm:equivariance}
Let $G_t \in \mathcal{G}$ be a noisy graph and $r \in SE(3)$ be a rigid transformation. The denoising network $\mathcal{M}_\theta$ satisfies (detailed proof in Appendix A):
\begin{equation}
    \mathcal{M}_\theta(r \circ G_t) = r \circ \mathcal{M}_\theta(G_t).
\end{equation}
\end{theorem}

\noindent\textbf{Remark.}
The theorem states that transforming the input EAG produces a correspondingly transformed network prediction. This claim applies specifically to the denoising network and does not assume strict
SE(3)-equivariance of the subsequent sampling and B-rep reconstruction operations.

\section{Experiments}
\label{sec:exp}

\subsection{Setup}
\label{subsec:exp_setup}

\subsubsection{Dataset Construction}
\label{subsubsec:dataset}

We introduce \textbf{EAGBench}, a comprehensive benchmark suite specifically designed for graph-based CAD generative modeling. EAGBench consists of two distinct datasets, \textit{EAGBench-DC} and \textit{EAGBench-FS}, enabling robust evaluation of generative performance across different data distributions and levels of modeling complexity.

We collect raw B-rep models from two major public repositories: DeepCAD~\cite{wu2021deepcad} and the Fusion 360 Gallery~\cite{willis2021fusion}. To ensure compatibility with our framework, we first perform geometric deduplication to remove redundant samples and further exclude models for which valid surface area or volume cannot be parsed. For each remaining model, we extract the hybrid attributes required by the Edge-Adjacency Graph. In particular, we parse the topology to ensure that each graph node corresponds to a unique topological edge in the B-rep. The geometric parameters ($\mathbf{p}_{\text{start}}, \mathbf{p}_{\text{end}}, \mathbf{p}_{\text{center}}$) are then extracted and normalized to the unit cube $[-1, 1]^3$. The resulting benchmarks exhibit diverse topological characteristics:

\textbf{EAGBench-DC}: Derived from the DeepCAD repository, this subset contains 95,392 unique models. It has an average of 16.6 nodes per graph and an average node degree of 3.7. The dataset is dominated by line segments, which account for 85.5\% of all primitives, reflecting the prevalence of prismatic features in standard mechanical components.

\textbf{EAGBench-FS}: Derived from the Fusion 360 Gallery, this subset contains 2,839 models. Compared with DeepCAD, these models exhibit greater geometric complexity. The dataset has an average node degree of 3.4, and its primitive distribution contains a higher proportion of curved geometries, including 14.3\% circles and 8.0\% arcs. This subset therefore serves as a challenging testbed for generating complex nonlinear structures.

\subsubsection{Baselines.}

We evaluate our framework against two categories of baselines to assess both topological correctness and geometric fidelity. For \textbf{graph generation}, we compare against state-of-the-art autoregressive and diffusion-based models, including GraphNVP~\cite{madhawa2019graphnvp}, GDSS~\cite{jo2022score}, ConGress~\cite{vignac2023digress}, and DiGress~\cite{vignac2023digress}, to validate the structural quality of the generated EAGs. For \textbf{CAD generation}, we benchmark against DeepCAD~\cite{wu2021deepcad}, a representative sequence-based method, as well as Text2CAD~\cite{khan2024text2cad} and TGBFN~\cite{zheng2025target}. We further compare with graph-based CAD generation methods, including BrepGiff~\cite{guo2025brepgiff} and GraphBRep~\cite{lai2026graphbrep}.

\begin{figure*}[t]
  \centering
  \includegraphics[width=\textwidth]{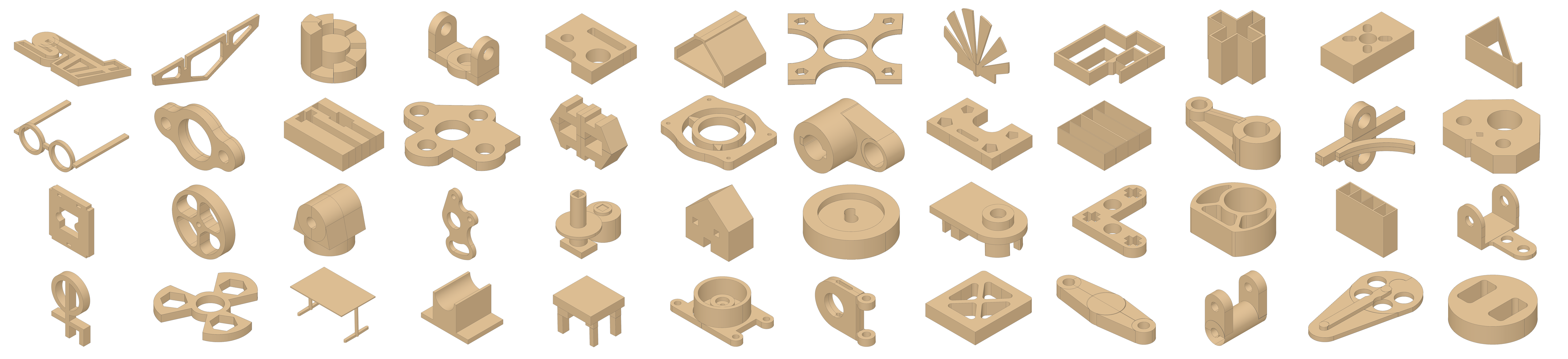}
  \vspace{-15pt} 
    \caption{\textit{Unconditional generation results on the EAGBench dataset.} GraphCAD generates realistic CAD models with complex geometry and topology.}
  \label{fig:vis_uncond}
\end{figure*}

\subsubsection{Metrics.}
We employ a comprehensive set of metrics covering both graph topology and geometric fidelity. For \textbf{graph structure}, following standard protocols~\cite{vignac2023digress}, we report Validity, Uniqueness, and Novelty, along with maximum mean discrepancy on degree distribution (\emph{Deg}), clustering coefficient (\emph{Clus}), and orbit counts (\emph{Orb}). For \textbf{CAD geometry}, we follow the evaluation protocol of NURBGen~\cite{usama2025nurbgen}. Specifically, we measure the Invalidity Ratio (\emph{IR}), as well as the squared Chamfer Distance (\emph{CD}) and the Hausdorff Distance (\emph{HD}), computed on $8{,}192$ uniformly sampled surface points after normalizing each shape to the $[-1,1]^3$ unit cube. To further assess distributional alignment and generative diversity, we report the Jensen-Shannon Divergence (\emph{JSD}) and the Minimum Matching Distance (\emph{MMD}). Note that Validity in Table~\ref{tab:graph_metrics} and IR in Table~\ref{tab:cad_metrics}, \ref{tab:ablation} are complementary measures computed using the same EAG-to-B-rep reconstruction pipeline.

\subsubsection{B-rep Reconstruction}
\label{app:surface_reconstruction}

Given a generated EAG $\mathcal{G}_0$, we instantiate lines, arcs, and circles using \textit{BRepBuilderAPI\_MakeEdge} and assemble adjacent edges into closed wires using \textit{BRepBuilderAPI\_MakeWire}. We then reconstruct compatible surfaces from these wires and stitch the resulting faces into a shell. Its watertightness is verified using \textit{BRepCheck\_Analyzer}, after which valid shells are converted into solids for uniform surface sampling. Sequence-based baselines use their official reconstruction implementations, whereas all graph-based methods share the pipeline above. The resulting B-reps are evaluated using identical metrics.

\begin{table*}[t]
\centering
\caption{Quantitative comparison of conditional generation. We evaluate reconstructed B-rep models in terms of topological validity (IR), geometric fidelity (CD, HD), and distributional quality (JSD, MMD). \textbf{Note:} For readability, CD, JSD, and MMD are each reported after scaling by $10^2$ within the $[-1, 1]^3$ unit cube.}
\vspace{-7pt}
\label{tab:cad_metrics}
\setlength{\tabcolsep}{4.5pt}
\begin{tabular}{l ccccc|ccccc}
\toprule
& \multicolumn{5}{c|}{\textbf{EAGBench-DC}} & \multicolumn{5}{c}{\textbf{EAGBench-FS}} \vspace{0.15cm}  \\
\textbf{Model} & \textbf{IR} $\downarrow$ & \textbf{CD} $\downarrow$ & \textbf{HD} $\downarrow$ & \textbf{JSD} $\downarrow$ & \textbf{MMD} $\downarrow$ & \textbf{IR} $\downarrow$ & \textbf{CD} $\downarrow$ & \textbf{HD} $\downarrow$ & \textbf{JSD} $\downarrow$ & \textbf{MMD} $\downarrow$ \\
\midrule
\multicolumn{2}{l}{\emph{Sequence-based}} \\
Transformer~\cite{vaswani2017attention} & 12.02 & 11.83 & 0.48 & 83.94 & 4.02 & 19.11 & 17.51 & 0.81 & 84.01 & 5.33 \\
DeepCAD~\cite{wu2021deepcad}  & 9.29 & 6.76 & 0.51 & 98.40 & 4.25 & 13.07 & 14.98 & 0.79 & 92.53 & 5.91 \\
Text2CAD~\cite{khan2024text2cad} & 3.87 & 3.36 & 0.45 & 59.63 & 3.98 & 8.29 & 4.62 & 0.63 & 70.38 & 4.78 \\
TGBFN~\cite{zheng2025target}    & \textbf{0.78} & 2.34 & 0.37 & 65.92 & 3.40 & 6.48 & 5.35 & 0.72 & 74.73 & 5.08 \vspace{0.15cm}  \\
\multicolumn{2}{l}{\emph{Graph-based}} \\
GraphBRep~\cite{lai2026graphbrep}    & 9.08 & 7.69 & 0.66 & 96.39 & 4.16 & 16.88 & 15.71 & 0.80 & 88.59 & 5.36 \\
BrepGiff~\cite{guo2025brepgiff}    & 2.17 & 2.27 & 0.39 & 67.23 & 3.62 & 6.53 & 4.72 & 0.74 & 73.91 & 4.29 \\
\rowcolor{gray!15}
\textbf{GraphCAD~(Ours)} & {1.67} & \textbf{1.73} & \textbf{0.28} & \textbf{33.22} & \textbf{1.97} & \textbf{5.42} & \textbf{3.37} & \textbf{0.59} & \textbf{64.97} & \textbf{3.52} \\
\bottomrule
\end{tabular}
\end{table*}

\begin{table*}[t]
\centering
\caption{
Ablation study on the EAGBench-DC dataset evaluating the contribution of individual components in GraphCAD.
We compare the full model with variants that remove the SE(3)-equivariant backbone (\emph{w/o SE(3) Equivariance}), semantic loss masking (\emph{w/o Loss Masking}), and point-based geometric rendering (\emph{w/o Geometric Rendering}).
}
\vspace{-7pt}
\label{tab:ablation}
\begin{tabular}{lcccccc|ccccc}
\toprule
& \multicolumn{6}{c|}{\textbf{Graph Metrics}} & \multicolumn{4}{c}{\textbf{CAD Metrics}} \vspace{0.15cm}  \\
\textbf{Model} & \textbf{Deg} $\downarrow$ & \textbf{Clus} $\downarrow$ & \textbf{Orb} $\downarrow$ & \textbf{Validity} $\uparrow$ & \textbf{Uniqueness} $\uparrow$ & \textbf{Novelty} $\uparrow$  & \textbf{CD} $\downarrow$ & \textbf{HD} $\downarrow$ & \textbf{JSD} $\downarrow$ & \textbf{MMD} $\downarrow$ \\
\midrule
\rowcolor{gray!15}
\textbf{Full Method~(Ours)} & \textbf{0.164} & \textbf{0.236} & \textbf{0.119} & \textbf{98.33} & \textbf{96.71} & \textbf{90.38} & \textbf{1.73} & \textbf{0.28} & \textbf{33.22} & \textbf{1.97} \\
w/o SE(3) Equivariance & 0.234 & 0.377 & 0.166 & 97.27 & 93.98 & 89.16  & 2.17 & 0.44 & 42.37 & 2.67 \\
w/o Loss Masking & 0.316 & 0.395 & 0.582 & 73.17 & 92.11 & 88.46 & 2.48  & 0.51 & 47.56 & 2.93 \\
w/o Geometric Rendering & 0.352 & 0.411 & 0.612 & 66.45 & 87.93 & 80.38 & 5.70 & 0.66 & 67.23 & 3.62 \\
\bottomrule
\end{tabular}
\end{table*}

\subsubsection{Implementation Details} 
\label{subsubsec:id} 
We implement a hybrid graph diffusion framework with $T=500$ timesteps. The SE(3)-equivariant \texttt{VNEncoder} processes local point sets via a three-stage hierarchy (32, 64, 128 channels) using VN-LeakyReLU activations. Rotation-invariant embeddings are derived via VN-Max-Pooling followed by an invariant linear projection. For diffusion, we utilize Markov transition matrices for discrete variables and a Gaussian variance schedule $\beta_t$ for continuous parameters. Training is conducted on 3 NVIDIA A800 GPUs. Please refer to Appendix C for additional configurations.

\subsubsection{Failure Mode Analysis of EAG-to-B-rep Reconstruction}
\label{app:failure_modes}

We categorize reconstruction failures into four primary types, as illustrated in Fig.~\ref{fig:failure_modes}: (i) \textbf{degenerate primitives} (Fig.~\ref{fig:failure_modes}a), where coincident endpoints ($\mathbf{p}_{\mathrm{start}}\approx\mathbf{p}_{\mathrm{end}}$) produce zero-length edges that cannot be incorporated into the B-rep wireframe; (ii) \textbf{undefined arc parametrization} (Fig.~\ref{fig:failure_modes}b), where collinear $\mathbf{p}_{\mathrm{start}}$, $\mathbf{p}_{\mathrm{end}}$, and $\mathbf{p}_{\mathrm{center}}$ fail to uniquely determine the plane and curvature of an arc or circle; (iii) \textbf{geometric inconsistency} (Fig.~\ref{fig:failure_modes}c), where the constraint $\lVert\mathbf{p}_{\mathrm{start}}-\mathbf{p}_{\mathrm{center}}\rVert \approx \lVert\mathbf{p}_{\mathrm{end}}-\mathbf{p}_{\mathrm{center}}\rVert$ is violated beyond the CAD kernel tolerance, resulting in invalid primitives; and (iv) \textbf{topological mismatch} (Fig.~\ref{fig:failure_modes}d), where graph-connected edges have non-coincident endpoints, producing an open wire that fails the watertightness check required for solid B-rep construction.

\subsection{Unconditional Topological Generation}
\label{subsec:uncond_topo}

We evaluate the ability of GraphCAD to capture the latent distribution of Edge-Adjacency Graphs (EAGs). As shown in Table~\ref{tab:graph_metrics}, GraphCAD consistently outperforms both flow-based (GraphNVP) and diffusion-based (GDSS, ConGress, DiGress) baselines across all evaluation metrics. In particular, our model achieves the lowest Deg, Clus, and Orb scores, indicating a closer match to the reference topological distribution. The advantage is especially pronounced on the more challenging EAGBench-FS benchmark: whereas the strongest baseline, DiGress, attains a validity of $77.45\%$, GraphCAD preserves a substantially higher validity of $96.58\%$. This large margin suggests that our geometry-aware formulation is particularly effective at modeling the intricate connectivity constraints inherent in engineering components.

Beyond structural validity, the consistently high Uniqueness ($>95\%$) and Novelty ($>92\%$) scores across both datasets indicate that GraphCAD produces diverse samples without evident overfitting or mode collapse. Qualitative results in Figure~\ref{fig:vis_uncond} further show that GraphCAD, trained on the graph-based EAGBench dataset, generates realistic and diverse CAD B-reps with complex topology and geometry. Together, these results validate that the EAG representation, coupled with our hybrid diffusion process, successfully grounds topological generation in physical geometric constraints, thereby promoting both structural soundness and sample diversity.

\subsection{Conditional CAD Generation}
\label{subsec:cond_gen}

We further evaluate the ability of GraphCAD to generate precise 3D CAD models under specific input constraints, following the protocol of TGBFN~\cite{zheng2025target}. As shown in Table~\ref{tab:cad_metrics}, our framework achieves a more favorable balance between topological validity and geometric accuracy. Although TGBFN attains a slightly lower IR on EAGBench-DC, GraphCAD consistently delivers superior geometric fidelity, reducing CD by approximately $26\%$ ($2.34$ to $1.73$).

On EAGBench-DC, GraphCAD achieves the best performance on all geometric and distributional metrics, attaining the lowest CD, HD, JSD, and MMD. In particular, compared with TGBFN, it reduces CD from $2.34$ to $1.73$ and substantially lowers JSD from $65.92$ to $33.22$. Although TGBFN achieves the lowest IR ($0.78\%$), our method remains highly competitive in validity ($1.67\%$) while providing clearly stronger geometric reconstruction quality.

\begin{figure}[t]
  \centering
  \includegraphics[width=\linewidth]{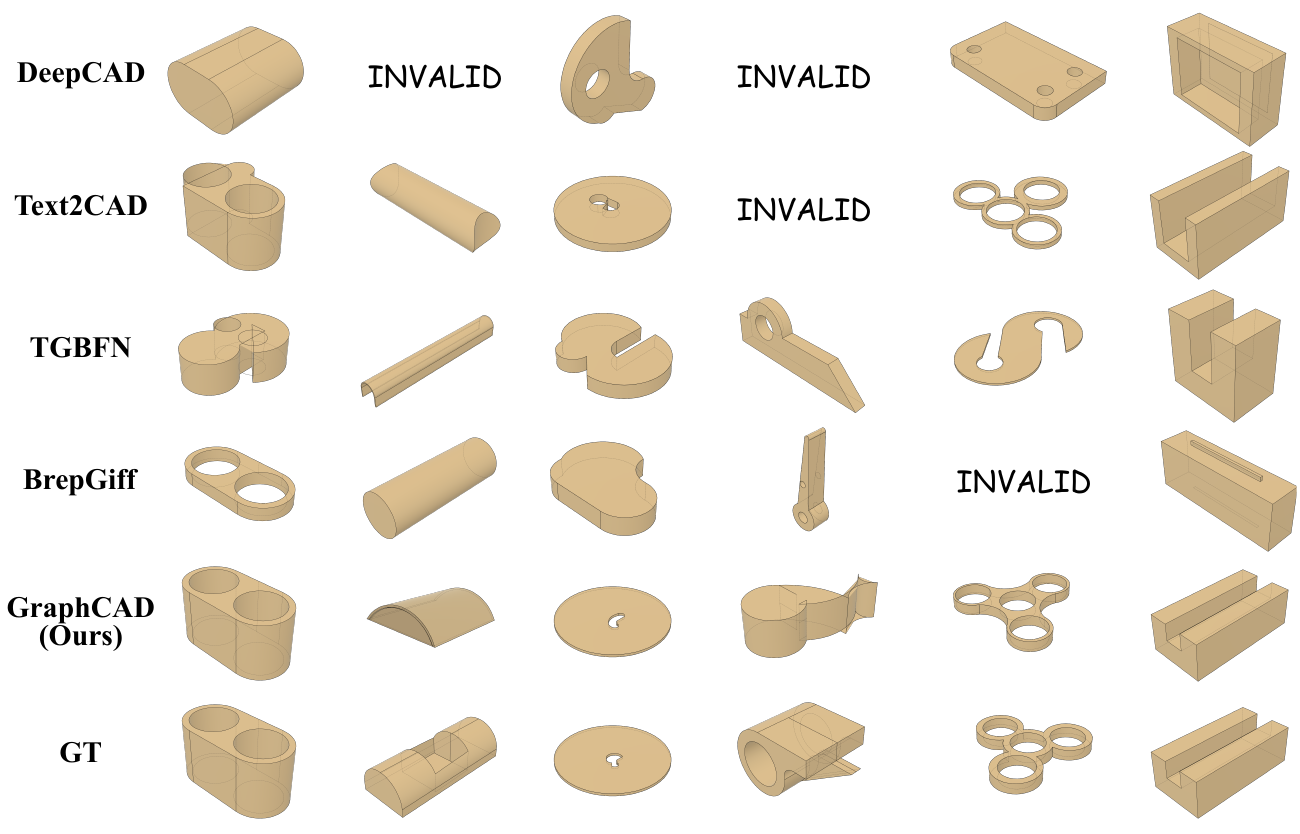}
  \vspace{-18pt}
    \caption{Qualitative comparison of CAD models from {conditional generation}. From top to bottom, we show results produced by DeepCAD, Text2CAD, TGBFN, BrepGiff, our proposed GraphCAD, and the Ground Truth (GT).}
  \label{fig:vis_cond}
  \vspace{-7pt}
\end{figure}

On the more challenging EAGBench-FS benchmark, GraphCAD consistently outperforms all baselines across all five metrics. Our method achieves the lowest IR ($5.42\%$), CD ($3.37$), HD ($0.59$), JSD ($64.97$), and MMD ($3.52$), surpassing both sequence-based methods, such as Text2CAD, and graph-based methods such as GraphBRep and BrepGiff. These results indicate that GraphCAD is not only more robust in preserving valid topology under complex conditions but also more accurate in recovering fine-grained geometry and the overall data distribution. The qualitative results in Figure~\ref{fig:vis_cond} further support this conclusion, showing that GraphCAD generates more coherent and geometrically faithful solids, whereas baseline methods more frequently exhibit structural or geometric artifacts.

\begin{table}[t]
    \centering
    \caption{\textit{Impact of point sampling density ($K$).} Trade-off between inference computational cost and generation quality on EAGBench-DC. Computational cost is measured on a single batch of 256 samples.}
    \label{tab:ablation_k}
    \vspace{-7pt}
    \resizebox{\linewidth}{!}{
    \begin{tabular}{c|cc|cc}
        \toprule
        \multirow{2}{*}{\textbf{Points ($K$)}} & \multicolumn{2}{c|}{\textbf{Computational Cost}} & \multicolumn{2}{c}{\textbf{Generation Quality}} \\
         & VRAM (GB) & Time (s) & IR ($\downarrow$) & CD ($\downarrow$) \\
        \midrule
        0  & 4.40 $\pm$ 0.03 & 131 $\pm$ 1 & 27.59 & 5.47 \\
        10 & 4.60 $\pm$ 0.02 & 145 $\pm$ 2 & 2.52 & 2.81 \\
        20 & 4.64 $\pm$ 0.02 & 146 $\pm$ 1 & 2.67 & 2.23 \\
        50 & 4.93 $\pm$ 0.01 & 149 $\pm$ 2 & 1.84 & 2.29 \\
        100 & 5.06 $\pm$ 0.03 & 151 $\pm$ 2 & 1.78 & 1.94 \\
        \bottomrule
    \end{tabular}
    }
    \vspace{-7pt}
\end{table}

\subsection{Ablation Study}
\label{subsec:ablation}

To validate our architectural design, we compare GraphCAD against variants lacking specific components. The results in Table~\ref{tab:ablation} confirm the necessity of each module. As shown in Table~\ref{tab:ablation}, the full GraphCAD model performs best overall, achieving the strongest balance between topological quality and geometric fidelity. Removing SE(3) equivariance leads to a moderate yet consistent decline across both graph and CAD metrics, with lower validity and worse reconstruction quality, indicating that equivariant geometric encoding improves robustness to spatial variation. Removing loss masking leads to a much larger drop in validity and noticeably weaker geometric consistency, confirming that suppressing supervision on undefined attributes is essential for stable learning. The most damaging ablation is removing geometric rendering, which causes the largest deterioration in validity, geometric fidelity, and distributional alignment, indicating that directly grounding node attributes in sampled spatial geometry is critical to the model. Overall, the ablation study shows that GraphCAD derives its advantage primarily from explicit geometry awareness, which tightly couples topological generation with faithful CAD reconstruction.

Table~\ref{tab:ablation_k} further examines the effect of the point sampling density $K$. Without explicit rendering ($K=0$), the model struggles to satisfy structural constraints, resulting in a high invalidity ratio of $27.59\%$. Introducing even a sparse point set ($K=10$) reduces this failure rate to $2.52\%$, underscoring the importance of spatial grounding for preserving topological consistency. Increasing $K$ to 100 further improves both geometric fidelity (CD decreases to $1.94$) and validity ($1.78\%$ IR), while incurring only a modest computational overhead. In particular, VRAM usage increases by $0.66$ GB (from $4.40$ GB to $5.06$ GB), supporting $K=100$ as a practical trade-off between reconstruction quality and efficiency.

\section{Limitations}

Several limitations remain. First, EAG models 1D boundary curves rather than explicit 2D surface patches. Although it supports curved boundaries and non-planar surfaces through line, arc, and circle primitives, complete surface definitions, such as NURBS control points, are recovered through post-processing. This reconstruction may introduce errors in regions with complex curvature. Second, the iterative diffusion process requires $T$ denoising steps, making inference slower than single-pass generative methods. Finally, our experiments focus on component-level parts. Scaling EAG generation to large assemblies with thousands of topological edges remains challenging because of the substantially larger discrete state space.

\section{Conclusion}

In this paper, we introduced GraphCAD, a geometry-aware hybrid graph diffusion framework for CAD generation built upon the proposed Edge-Adjacency Graph (EAG) representation. By combining a curve rendering mechanism with an SE(3)-equivariant backbone, our model grounds abstract node attributes in spatial geometry, thereby promoting robust topological and geometric consistency. Extensive experiments on the EAGBench benchmark demonstrate that GraphCAD achieves state-of-the-art performance in both validity and fidelity compared with existing sequence-based and graph-based baselines. In future work, we plan to extend this probabilistic formulation to infer full surface definitions, moving beyond wireframe generation toward complete solid reconstruction.

\begin{acks}
This work is supported by the National Key R\&D Program of China under Grant 2024YFB3312503, the National Major Scientific Instruments and Equipments Development Project of National Natural Science Foundation of China under Grant 62427820, the Science Fund for Creative Research Groups of Sichuan Province Natural Science Foundation under Grant 2024NSFTD0035, the Ministry of Education Engineering Research Center Guiding Project for Machine Learning and Industrial Intelligence Applications under Grant SCU2024D013, the Science and Technology Major Project of Sichuan Province under Grant 2024ZDZX0003 and 2025ZDZX0140. We also acknowledge the support of the Sichuan Province Engineering Technology Research Center of Broadband Electronics Intelligent Manufacturing.  
\end{acks}

\bibliographystyle{ACM-Reference-Format}
\balance
\bibliography{sample-base}










\end{document}